\documentclass[sigconf,nonacm]{acmart}
\renewcommand\footnotetextcopyrightpermission[1]{} 
\usepackage{subfigure}
\AtBeginDocument{%
  }

\begin{document}

\title{MAO: A Framework for Process Model Generation with Multi-Agent Orchestration}

\author{Leilei Lin}
\affiliation{%
  \institution{Capital Normal University}
  \city{Beijing}
  \country{China}}
\email{leilei_lin@126.com}

\author{Yumeng Jin}
\affiliation{%
  \institution{Capital Normal University}
  \city{Beijing}
  \country{China}}
\email{jinyumeng2012@163.com}

\author{Yingming Zhou}
\affiliation{%
 \institution{Capital Normal University}
 \city{Beijing}
 \country{China}}
 \email{aizhouym@gmail.com}

\author{Wenlong Chen}
\affiliation{%
  \institution{Capital Normal University}
  \city{Beijing}
  \country{China}}
  \email{wenlongchen@sina.com}

\author{Chen Qian}
\authornotemark[1]
\affiliation{%
  \institution{Tsinghua University}
  \city{Beijing}
  \country{China}}
  \email{qianc62@gmail.com}

\begin{abstract}
Process models are frequently used in software engineering to describe business requirements, guide software testing and control system improvement. However, traditional process modeling methods often require the participation of numerous experts, which is expensive and time-consuming. Therefore, the exploration of a more efficient and cost-effective automated modeling method has emerged as a focal point in current research. This article explores a framework for automatically generating process models with \textbf{multi-agent orchestration} (MAO), aiming to enhance the efficiency of process modeling and offer valuable insights for domain experts. Our framework MAO leverages large language models as the cornerstone for multi-agent, employing an innovative prompt strategy to ensure efficient collaboration among multi-agent. Specifically, 1) generation. The first phase of MAO is to generate a slightly rough process model from the text description; 2) refinement. The agents would continuously refine the initial process model through multiple rounds of dialogue; 3) reviewing. Large language models are prone to hallucination phenomena among multi-turn dialogues, so the agents need to review and repair semantic hallucinations in process models; 4) testing. The representation of process models is diverse. Consequently, the agents utilize external tools to test whether the generated process model contains format errors, namely format hallucinations, and then adjust the process model to conform to the output paradigm. The experiments demonstrate that the process models generated by our framework outperform existing methods and surpass manual modeling by 89\%, 61\%, 52\%, and 75\% on four different datasets, respectively.
\end{abstract}



\maketitle

\section{Introduction}

Software engineering is a field that employs a systematic and disciplined methodology for the design, implementation, and maintenance of software systems \cite{biolchini2005systematic}. Process-driven approaches \cite{schaffer2021process} are commonly utilized within software engineering, focusing on process modeling to aid software developers in gaining a comprehensive understanding of software systems, thereby enhancing development efficiency or optimizing running software systems \cite{drave2022model}. For example, a process model acts as a guide or blueprint for developers throughout software development \cite{corradini2018guidelines}, it facilitates communication between stakeholders in the company and developers during the requirements analysis phase \cite{grolinger2014integration}, and it is a valuable auxiliary tool for software testing \cite{von2022approach}. However, over the past few decades, although traditional manual modeling approaches have yielded significant benefits, their inherent drawbacks such as time-consuming processes and high costs have become increasingly evident.
\begin{figure*}[t]
  \centering
  \includegraphics[width=0.98\textwidth]{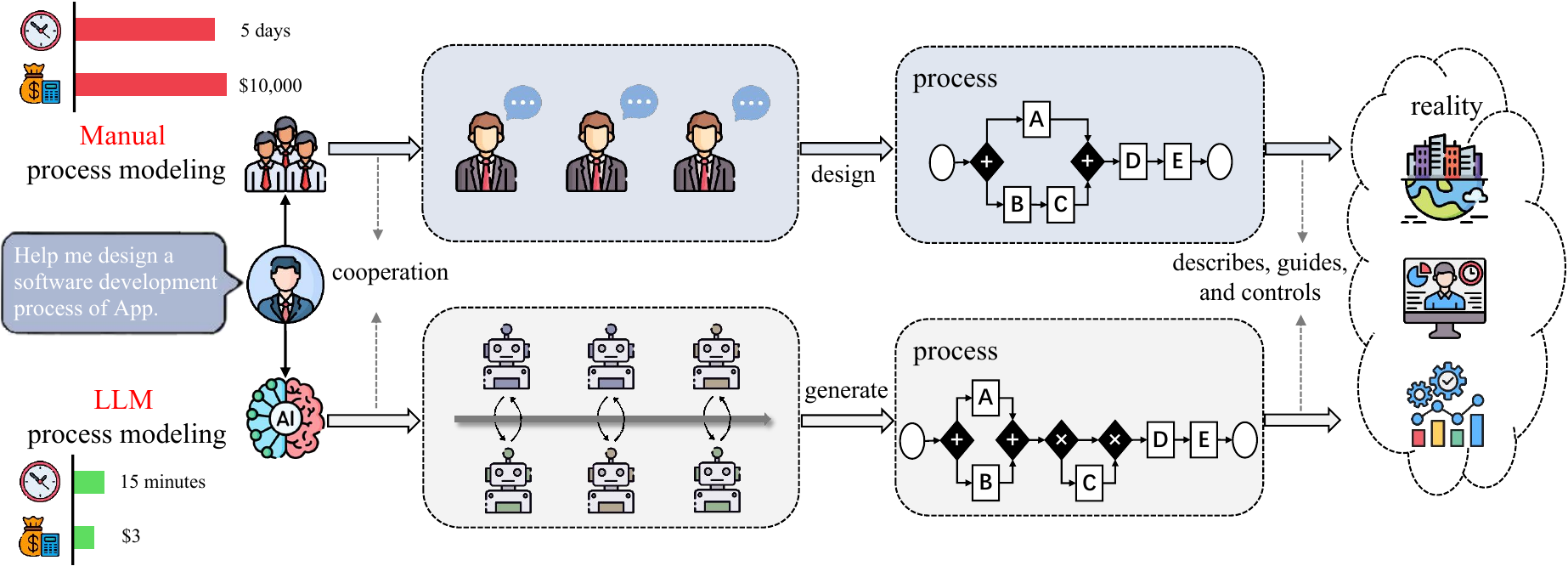}
  \caption{Traditional process modeling is time-consuming and costly, whereas LLM-based multi-agent can efficiently generate process models.}
  \label{fig:introduction}
\end{figure*}

This motivated academics to devise methods and tools that enable business analysts or software developers to create process models as efficiently as possible. Process mining, as we all know, is an automated process modeling technique that uses event logs from information systems to reconstruct process models and improve ongoing business processes \cite{huser2012process}. However, process mining belongs to a data-driven approach, relying on predefined mining rules by experts, resulting in significant disparities in process models generated by different rules \cite{lekic2016discovering, guo2015mining, sommers2021process}. With the rapid advancement of large language models (LLMs), LLMs are emerging with intelligence that was previously unique to human beings \cite{wei2022emergent}. Specifically, the agents based on LLMs have attracted significant attention and exhibited a certain level of human-like intelligence, such as the ability to use tools \cite{schick2024toolformer}, play games \cite{wang2023voyager, chen2023agentverse}, and develop software \cite{qian2023communicative}. Therefore, a meaningful idea naturally arises: could LLM-based agents be employed for process modeling to promote future progress in the software engineering domain?

Some researches about workflow automation generation based on large language models has shown initial progress in the field of Robotic Process Automation (RPA), like ProAgent \cite{ye2023proagent} and FlowMind \cite{zeng2023flowmind}. However, these researches mainly focus on generating activities to be executed and arranging them in a sequential order, without considering complex relationships such as concurrency or choice between activities. It is important to note that sorting through the complex relationships among activities is the most difficult part of process modeling. Kourani et al. \cite{kourani2024process} proposed the ProMoAI method, which involves converting process text into Partially Ordered Workflow Language (POWL) \cite{kourani2023powl} before extracting the process model. This approach relies on the capabilities of POWL and does not fully leverage the capabilities of LLMs. Furthermore, ProMoAI requires user involvement in resolving hallucination phenomena during the process generation.

In this paper, we explore a framework for process model generation with LLM-based multi-agent to overcome the above challenges. As illustrated in Figure \ref{fig:introduction}, traditional process modeling often involves multiple individuals engaging in discussions to design the final process model upon receiving user requirements. Similarly, in our framework MAO, different agents will also play different roles that collaborate through four carefully orchestrated phases (i.e., Generation, Refinement, Reviewing, Testing) to collectively complete the process model generation. The process model in this article is represented using business process model and notation (BPMN), a notation standardized by the Object Management Group (OMG) \cite{specification2006object}. As shown in Figure \ref{fig:introduction}, rectangular boxes represent activities: language choice (A), demand analysis (B), design review (C), coding (D), testing (E). Diamond boxes represent gateways, where the parallel gateway + denotes a concurrency relationship, and the exclusive gateway x denotes a choice relationship. Circles represent the start and end of the process. Additionally, to enhance agents' understanding of BPMN, a BPMN text format based on XML structure is designed. To improve agents' performance during the process, we integrate context learning (e.g., few-shot) \cite{liu2022few} and Chain of Thought (CoT) \cite{wei2022chain} mechanisms into the prompting process. Furthermore, when addressing hallucination phenomena, agents autonomously undertake the repair process without the need for manual intervention.

\noindent
In summary, our contributions are outlined as follows:

\begin{itemize}
\item[-] We propose MAO, a novel framework that allows many agents to collaborate to automatically generate process models from textual process requirements without the manual involvement.
\item[-] We leverage agents' inherent abilities to autonomously rectify hallucination phenomena. In addressing hallucinations within process models, we categorize them into semantic and format hallucinations, where the former are resolved through multi-round interactions among agents, and the latter involve the integration of external tools for effectively handling format errors in the process models.
\item[-] Experimental results on public datasets demonstrate that our framework MAO effectively harnesses the potential of large language models to generate useful process models for users. Furthermore, the quality of process models generated by MAO outperforms existing baselines and surpasses the level of manual modeling on four publicly available datasets.  
\end{itemize}




Next, we will introduce preliminary knowledge in section 2,  the proposed framework MAO in section 3, and  section 4 will discuss the experiments results and analysis, as well as relevant literature in section 5.

\section{Preliminary}
\subsection{Process Model BPMN}
A process model can be considered as a tool or language utilized for modeling software systems or service requirements. For instance, the Web Services Business Process Execution Language (WS-BPEL) serves as an XML-based implementation-oriented standard for delineating the interaction of executable processes with web services,  without providing a graphical notation \cite{ings2010ws}. In contrast, the event-driven process chain (EPC) was introduced with the objective of semantically modeling business processes through the identification and graphical documentation of business management interconnections within an organization \cite{amjad2018event}. With the aim of graphical software specification, design, and documentation, the Object Management Group (OMG) introduced the unified modeling language (UML), primarily utilized for the initial specification of IT systems \cite{specificationuml}. Furthermore, the Petri net is a mathematical modeling tool \cite{peterson1981petri, he2022petri, van2000verification} commonly employed to model and analyze the behavior of systems like manufacturing processes, communication protocols, and computer networks. Petri nets are particularly adept at capturing issues of concurrency, synchronization, and resource allocation within intricate systems. However, one minor limitation of Petri nets is their propensity to generate overly complex models, rendering them challenging for modelers to comprehend. This inherent issue stems from the fact that Petri nets consist solely of places and transitions as their two fundamental elements. Conversely, the BPMN standard terminology has rich modeling elements that can help modelers quickly generate process models for complex tasks \cite{freund2012real}. What is more, compared to EPC, the graphical notation of BPMN facilitates the execution of modeled processes within a process engine \cite{oukharijane2019towards, gonzalez2017business}.

\begin{table}
  \caption{The BPMN elements and descriptions.}
  \label{tab:table}
  \begin{tabular}{ccl}
    \begin{minipage}[h]{0.97\columnwidth}
    \centering
    {\includegraphics[width=\textwidth]{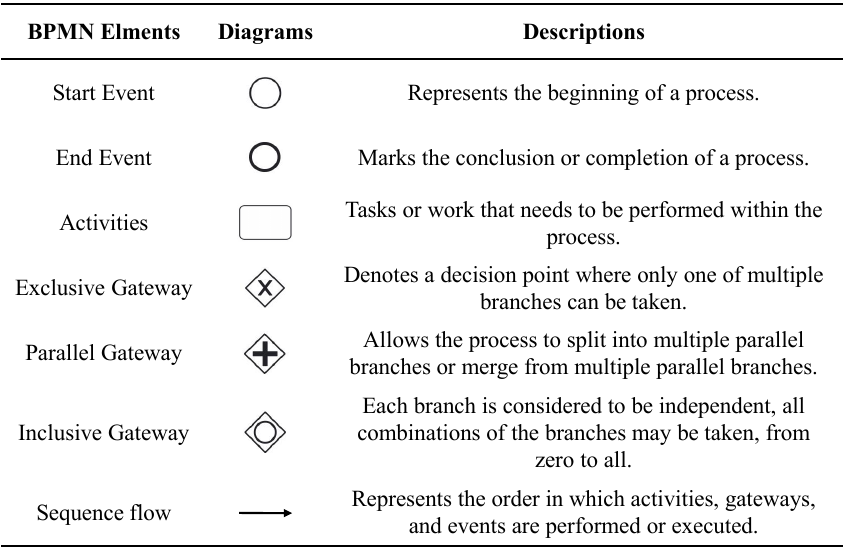}}
  \end{minipage}
\end{tabular}
\end{table}

Based on the above discussion, we select BPMN as the expression form for process models in this paper. It is important to note that BPMN contains a rich set of modeling elements (such as data flow, message flow, human interaction packages, etc.), but this paper only supports common types of elements as presented in the Table \ref{tab:table}. The reason for this limitation is that the current capabilities of large language models are not yet sufficient to support such a wide range of modeling elements (as discussed in the experimental section). On the other hand, from the perspective of sequence flow, the core elements outlined in Table \ref{tab:table} generally meet the most of the needs of modelers.

\subsection{BPMN Text for LLMs}
\begin{figure}[t]
  \centering
  \includegraphics[width=0.48\textwidth]{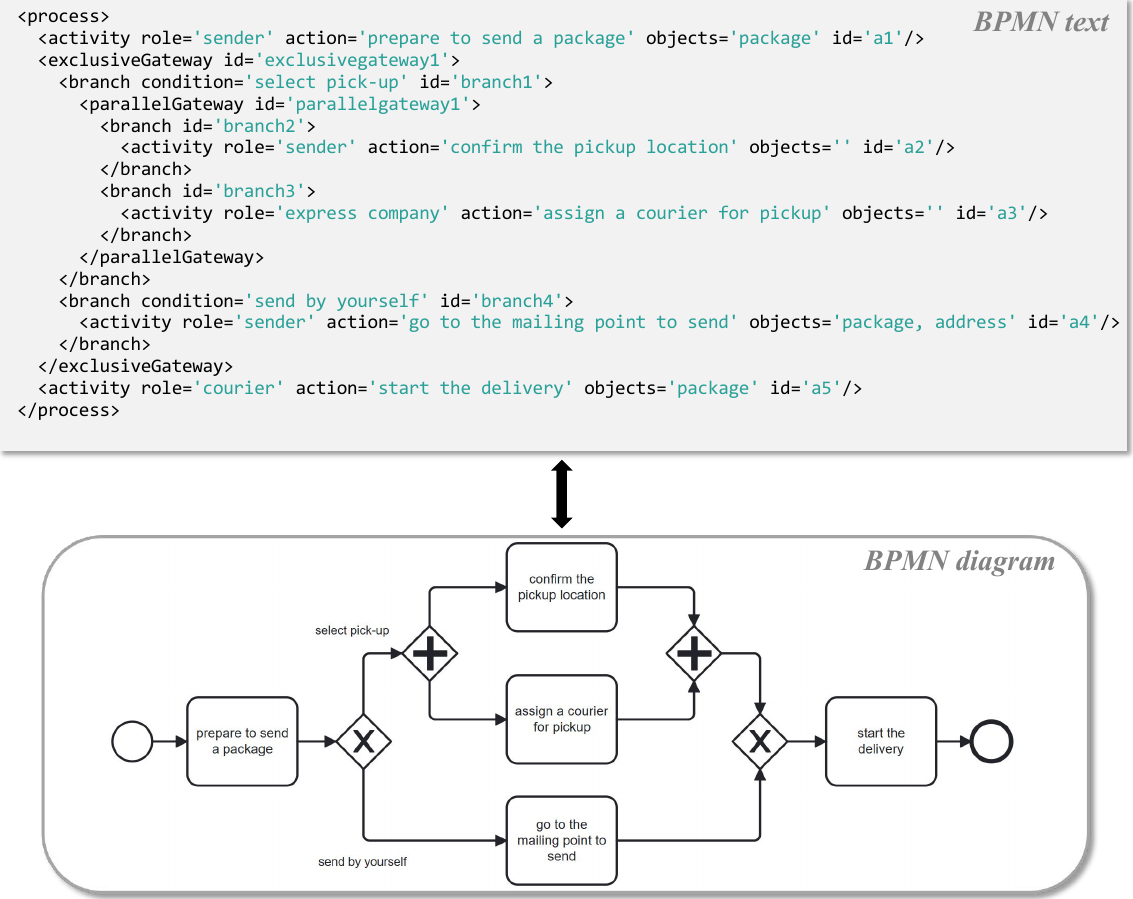}
  \caption{BPMN text and BPMN diagram.}
  \label{fig:case}
\end{figure}

As we know, each BPMN diagram is represented by a .bpmn file in real-life, but this file contains abundant content. In order to enhance the understanding of BPMN diagrams by LLMs, we design a more concise BPMN text for the agent's operation. As shown in Figure \ref{fig:case}, on the bottom side is a simple BPMN diagram, while on the top side is its equivalent BPMN text representation. In BPMN text, <process> signifies the beginning of a process model, while </process> indicates the end. The activity node <activity> includes four attributes: the executor ``role'', the activity name ``action'', the resources ``object'' required for the activity, and a unique identifier ``id'' for the activity, where the value of the ``object'' attribute can be null. The branch node <branch> must be enclosed within the gateways like exclusiveGateway, parallelGateway and inclusiveGateway, where the ``condition'' attribute in <branch> denotes the condition that needs to be met for the branch to be executed. In this paper, regardless of whether it is generating a process model or modifying a process model, multi-agent operate on the BPMN text.

\section{Process Modeling Framework}
\begin{figure*}[h]
  \centering
  \includegraphics[width=\textwidth]{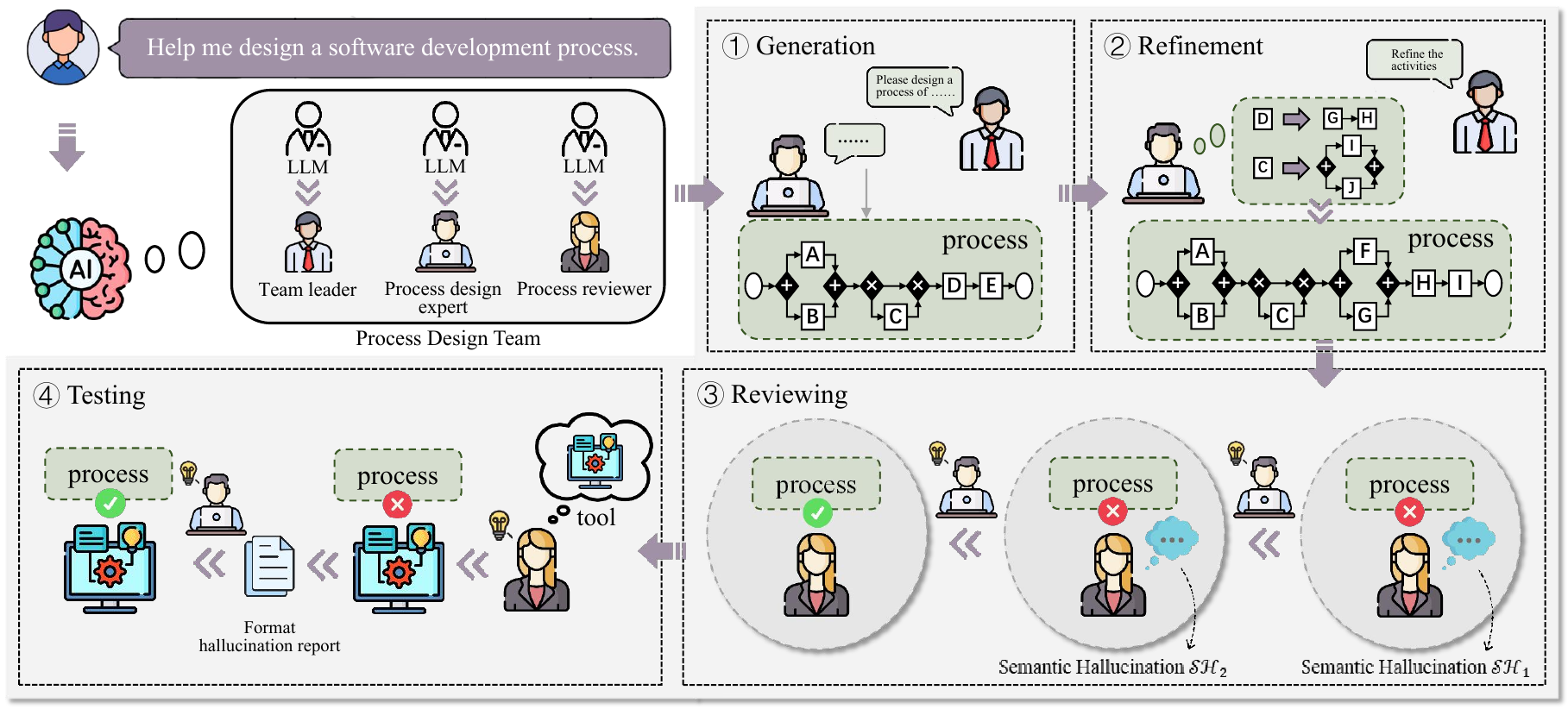}
  \caption{Overview of our framework named MAO for process modeling.}
  \label{fig:overview}
\end{figure*}

In this section, a novel framework (i.e., MAO) will be presented, which enables the generation of process models through multi-agent orchestration. Figure \ref{fig:overview} illustrates the specifics of MAO. Upon receiving modeling requirements from users, we first establish a process design team comprised of multi-agent with distinct roles who collaborate to complete the process modeling tasks. The team collaboration involves four main phases: the first phase involves the team leader and process design expert in generating the initial process model; the second phase entails further refinement based on the initial model; the third phase involves the process reviewer and process design expert adjusting semantic hallucinations within the process model; and in the final phase, external tools are employed to rectify any format errors in the process model.

\begin{figure}[h]
  \centering
  \includegraphics[width=0.48\textwidth]{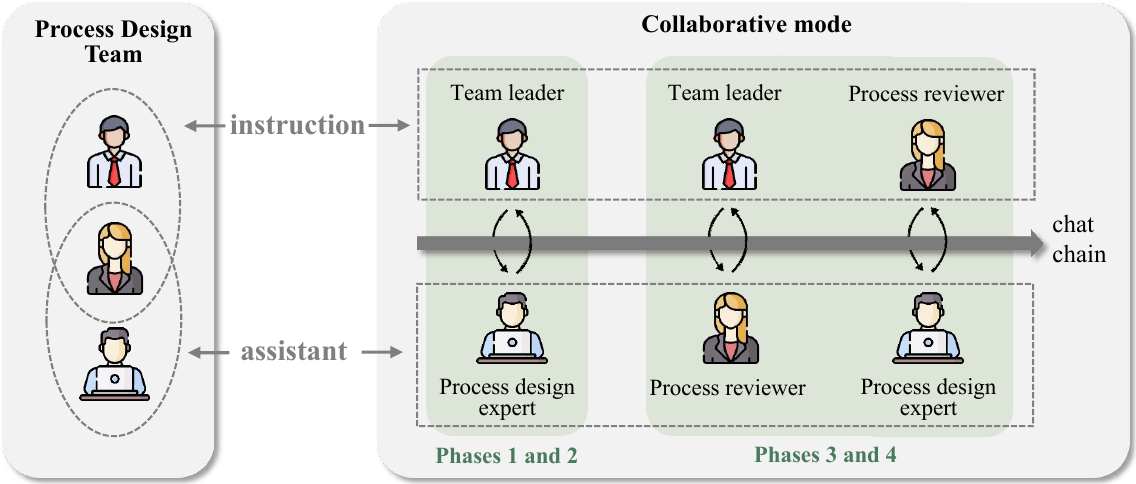}
  \caption{Collaborative mode of the multi-agent in MAO.}
  \label{fig:chatchain}
\end{figure}

It is worth noting that in the establishment of the process design team, we employ inception prompting \cite{li2023camel}, a method that has demonstrated effectiveness in enabling agents to fulfill their respective roles. As illustrated in Figure \ref{fig:chatchain}, we categorize the roles into two types: instructor and assistant, with the former responsible for thinking and issuing directives, and the latter for executing specific tasks based on the directives received. Within the MAO framework, the first two phases involve only the team leader and process design expert, while in the third and fourth phases, all three roles are involved.

\subsection{Generation}
\begin{figure}[t]
  \centering
  \includegraphics[width=0.48\textwidth]{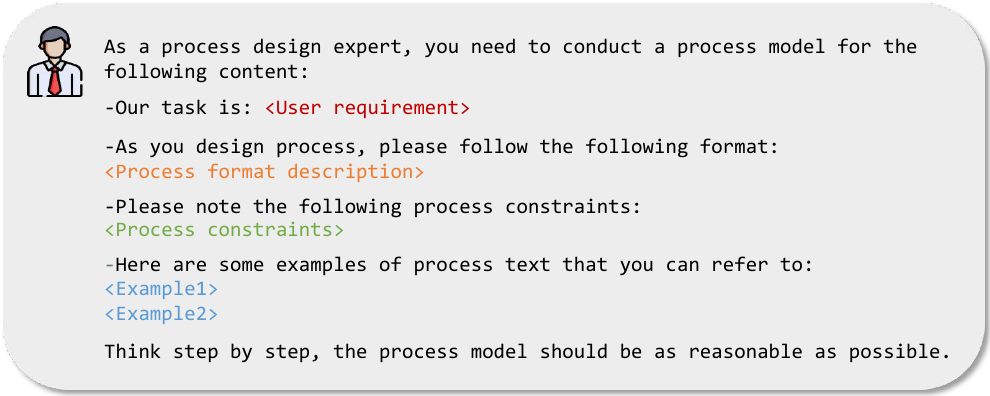}
  \caption{The example for prompting the agents to generate BPMN.}
  \label{fig:phase1prompt}
\end{figure}

The first phase of MAO involves generating an initial process model from user requirements. Initially, the team leader creates instruction content based on user needs, and then guides the process design expert in completing the process modeling through prompt engineering. In order to better stimulate the capabilities of LLMs, the following prompting strategies have been adopted:

\begin{itemize}

\item \textit{Knowledge Injection}. This strategy entails providing the LLMs with novel, specific information or context that may not have been part of its original training \cite{martino2023knowledge}. Although a brief introduction to BPMN content is provided to each agent during role assignment, it is deemed necessary to further inject more specific BPMN knowledge to the agents during the actual generation process. As depicted in Figure \ref{fig:formatprompt}, the injected knowledge comprises two parts: i) process format description. We provide the agents with the BPMN text format, which includes <process>, <activity>, <exclusiveGateway>, and so on, in order to take advantage of LLMs capabilities in generating BPMN text; ii) process constraints. We list a few constraints to help the BPMN text generated by the agents follow these constraints as closely as possible, hence improving the quality of BPMN text. For instance, C\_2 emphasizes that a parallel gateway must include two branches. Naturally, these constraints can be dynamically expanded based on actual requirements.

\begin{figure}[b]
  \centering
  \includegraphics[width=0.48\textwidth]{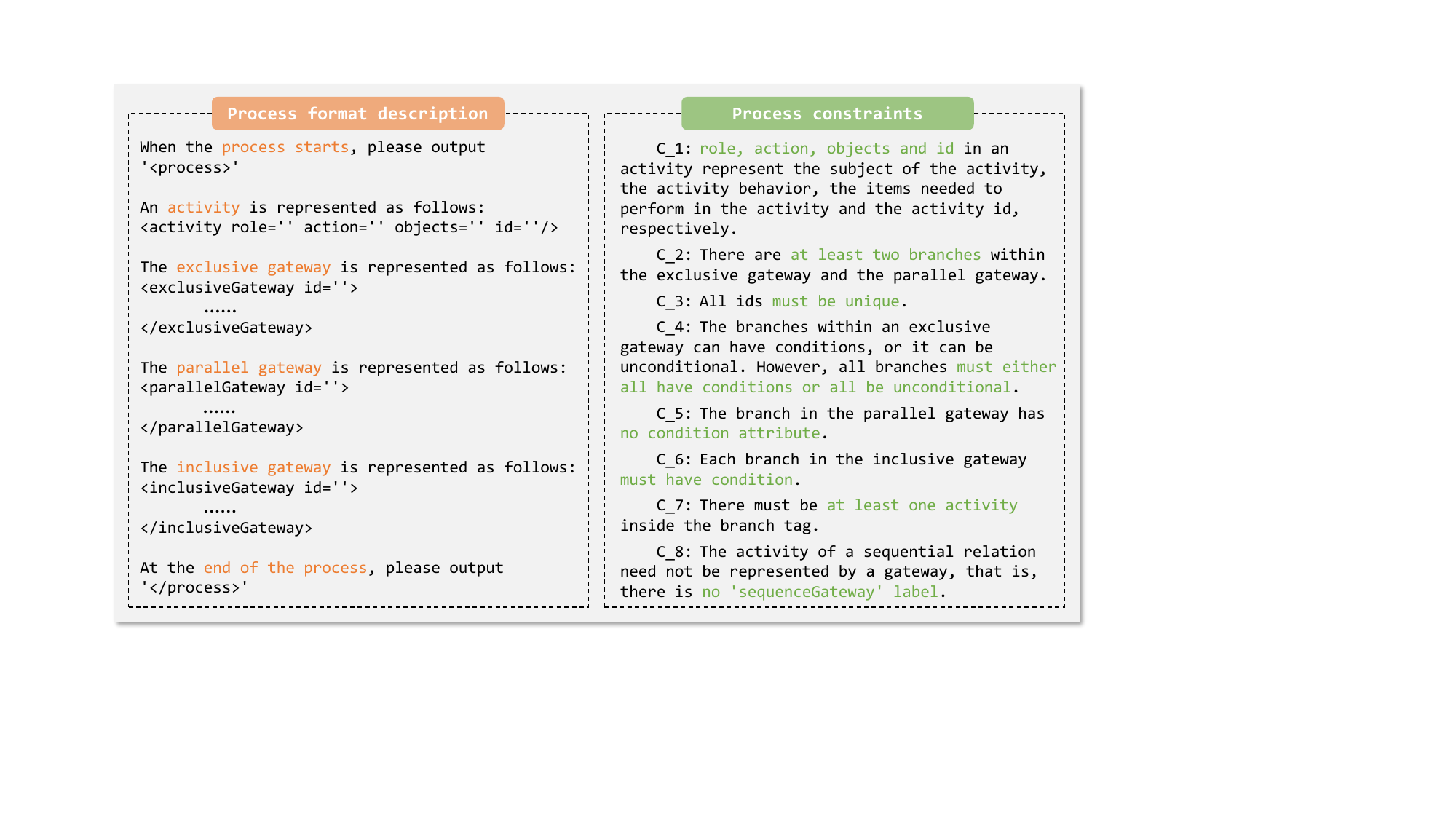}
  \caption{Two parts for knowledge injection.}
  \label{fig:formatprompt}
\end{figure}

\item \textit{Few shot learning}. This approach entails instructing the agents on solving the task by presenting a few examples of input and expected output \cite{brown2020language}. So, we provide the examples of BPMN text to the agents. For instance, the BPMN text presented in Figure \ref{fig:case} serves as an illustrative example.

\item \textit{Chain-of-thought}. The concept of Chain-of-Thought (CoT) has been proposed to enhance the capacity of large language models in handling more intricate tasks, with its fundamental principle being ``let's think step by step'' \cite{wei2022chain, kojima2022large}. Inspired by this notion, we also incorporate the CoT approach into our prompting strategy to reinforce the agent's generation of process models.

\end{itemize}
Figure \ref{fig:phase1prompt} shows the overall content of our prompt strategy. When guiding the process design expert to generate a process model (i.e., BPMN text), the team leader would include user requirements and examples in the instructions, as well as process format descriptions and process constraints.

\subsection{Refinement}
After establishing the initial process model, the agent of the team leader proposes instructions for refining the activities in the model, then the process design expert autonomously decides on the specifics to be refined based on the instructions. The orchestration at this phase is primarily aimed at addressing two causes that can easily lead to low quality process models: (1) overly vague user requirements. The ideal user requirement involves providing a detailed description of the process, including the activities to be modeled and the dependencies between activities. However, in reality, many users only present a modeling goal without any description of the process content, such as designing a procurement process. This vague requirement can result in the process model generated by the agent being very rough, possibly with only a few common activities. We know that a very coarse process model of procurement process may lack sufficient value as a reference for users; (2) low information density in LLMs. Large language models are typically trained on vast amounts of unlabeled or weakly labeled text data \cite{smith2024language, naveed2023comprehensive}, which may encompass diverse styles and themes, resulting in low information density. As a result, when generating responses, LLMs may opt for a more neutral and generalized expression to cater to various users. 

The refinement operation mainly consists of two actions: refining existing activities into multiple sub-activities and adding appropriate gateways. The team leader provides a prompt including the initial model generated by the first phase and user requirements to the process design expert, enabling the expert to autonomously decide on the content to be refined. The prompts also include a few examples of refinement operations to guide the expert's refinement actions.

\subsection{Reviewing}
\begin{table}
  \caption{Categories of semantic hallucinations, and the 
illustrated examples from the delivery process in Figure \ref{fig:case}.}
  \label{tab:table2}
  \begin{tabular}{ccl}
    \begin{minipage}[h]{0.97\columnwidth}
    \centering
    {\includegraphics[width=\textwidth]{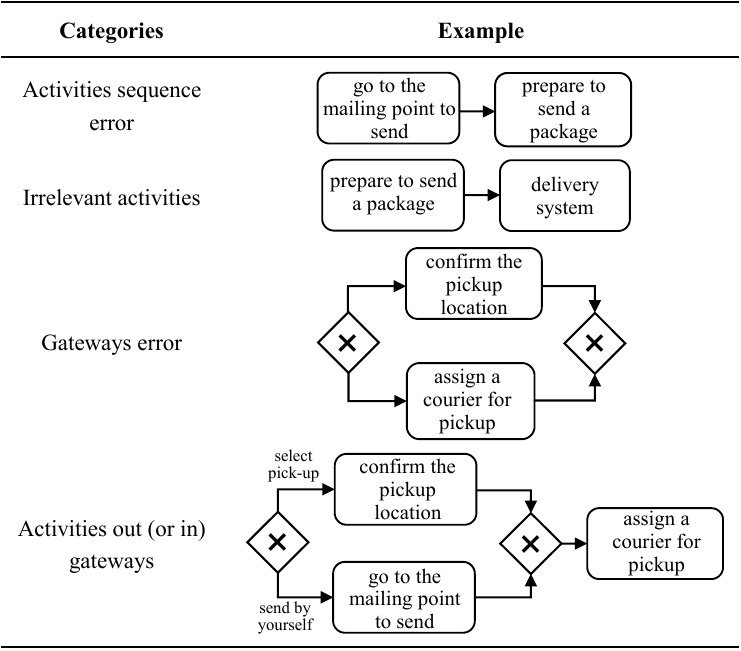}}
  \end{minipage}
\end{tabular}
\end{table}
In this section, we will discuss how multi-agent collaborate to address hallucination phenomena in process modeling. In natural language processing, ``hallucination'' denotes the situation where a machine learning model generates text or responses that are entirely disconnected from the context or not reliant on the input data \cite{huang2023survey}. This can manifest when the agents produce language that lacks logic in relation to the input or sounds plausible but lacks factual accuracy. This paper attributes illogical errors in process modeling to semantic hallucinations ($\mathcal{SH}$), in other words, the focus of the Reviewing phase in MAO is on rectifying logical errors in process models. As depicted in Table \ref{tab:table2}, we present the categories of semantic hallucinations along with corresponding examples: i) the first common logical error involves activities occurring out of sequence. For example, the activity ``prepare to send a package'' must be executed before ``go to the mailing point to send'' in the delivery process as shown in Figure \ref{fig:case}; ii) the second type of logical error involves irrelevant activities appearing in the process model. The activity ``delivery system'' may seem related to the delivery business, but it does not contribute to process modeling; iii) the third is gateways error, where activities ``confirm the pickup location'' and ``assign a courier for pickup'' must both be executed, but the results generated by agents indicate the selection of both activities, which clearly contradicts reality; iv) the fourth error is that the activities must be inside or outside the gateway branch. For example, the activities ``assign a courier for pickup'' and ``go to the mailing point to send'' are unrelated and ``assign a courier for pickup'' must appear in the upper branch of the exclusive gateway, as shown in Figure \ref{fig:case}.

When addressing semantic hallucinations, three roles are involved in the process: first, the team leader prompt the process reviewer to identify semantic hallucinations in BPMN text, where the prompt includes categories and examples of semantic hallucinations; second, the reviewer identifies semantic hallucinations and generate corresponding modification suggestions to alert the process design expert; finally, the expert makes modifications to the BPMN text based on the reviewer's suggestions. It is important to note that the entire process of addressing semantic hallucinations is repeatable.

\subsection{Testing}
After processing semantic hallucinations, we will make the final formatting adjustments to the BPMN text. Although in the first phase of MAO, we enhance the agents' understanding of the BPMN text format through knowledge injection, the agents still make a few errors when generating complex process models, such as missing attribute values, etc. In addition, the agents also make format errors during Refinement and Reviewing phases. Therefore, we collectively refer to content that does not conform to process format description and process constraints as format hallucinations.

\begin{figure}[h]
  \centering
  \includegraphics[width=0.48\textwidth]{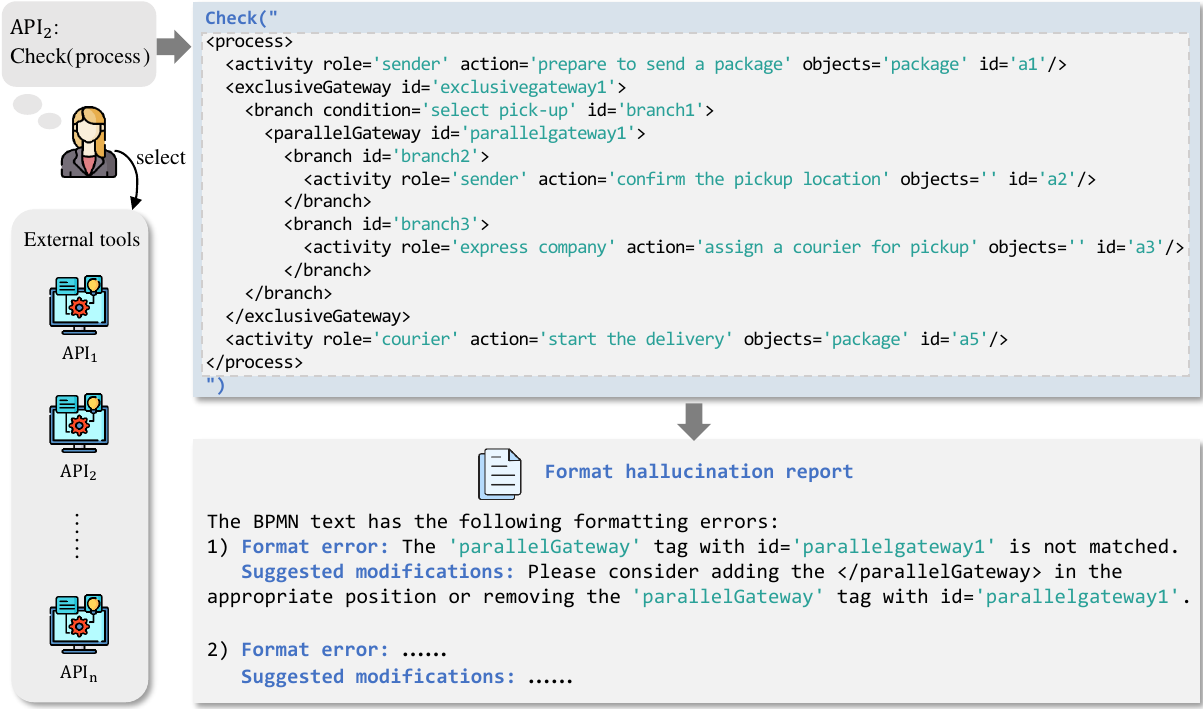}
  \caption{An example for illustrating the format hallucination report.}
  \label{fig:test}
\end{figure}

Large language models achieve impressive few-shot results on a variety of natural language processing tasks \cite{brown2020language, chowdhery2023palm}, but all of these models have several inherent limitations like difficulties in understanding low-resource languages \cite{lin2021few} and a lack of mathematical skills to perform precise calculations \cite{patel2021nlp}. Thus, we are preparing to use external tools to enhance agents' ability to edit format hallucinations, while the use of external tools has been proven effective in dealing with such issues \cite{schick2024toolformer}. The benefits of using external tools for positioning format hallucinations are: (a) high efficiency. If the process model is too complex, using agents to detect format hallucinations can easily lead to omissions and misjudgments. External tools can accurately locate format errors in BPMN text through simple traversal and calculations; (b) high scalability. The agents use external tools in the form of API calls, so it can meet diverse format error checking needs without affecting the MAO framework. The process of handling format hallucinations mainly involves three steps. First, the team leader would inform the process reviewer of which external tools' APIs can be used. Then, the reviewer selects the external API to check for format errors in the BPMN text and generate a report to prompt the process design expert. Finally, the expert will modify the BPMN text based on the report content. Figure \ref{fig:test} illustrates the process of handling format hallucinations and the content of the report with an example. In addition to identifying the location of the errors, the report also provides suggestions for modifications.

\section{Experiment}
In this section, we conduct an evaluation of the performance of the MAO framework. All our experiments are conducted using GPT-4, and all the code is publicly available\footnote{\url{https://anonymous.4open.science/r/MAO-1074}}.

\subsection{Setup}
\begin{table}
  \caption{The four processes from FG-C along with the corresponding manual modeling information statistics, where \textit{Model number} represents the BPMN models drawn by different individuals, and \textit{Distance} indicates the disparity between the manually modeled BPMN and the standard answers provided in the dataset.}
  \label{tab:e_table1}
  \begin{tabular}{ccl}
    \begin{minipage}[h]{0.96\columnwidth}
    \centering
    {\includegraphics[width=\textwidth]{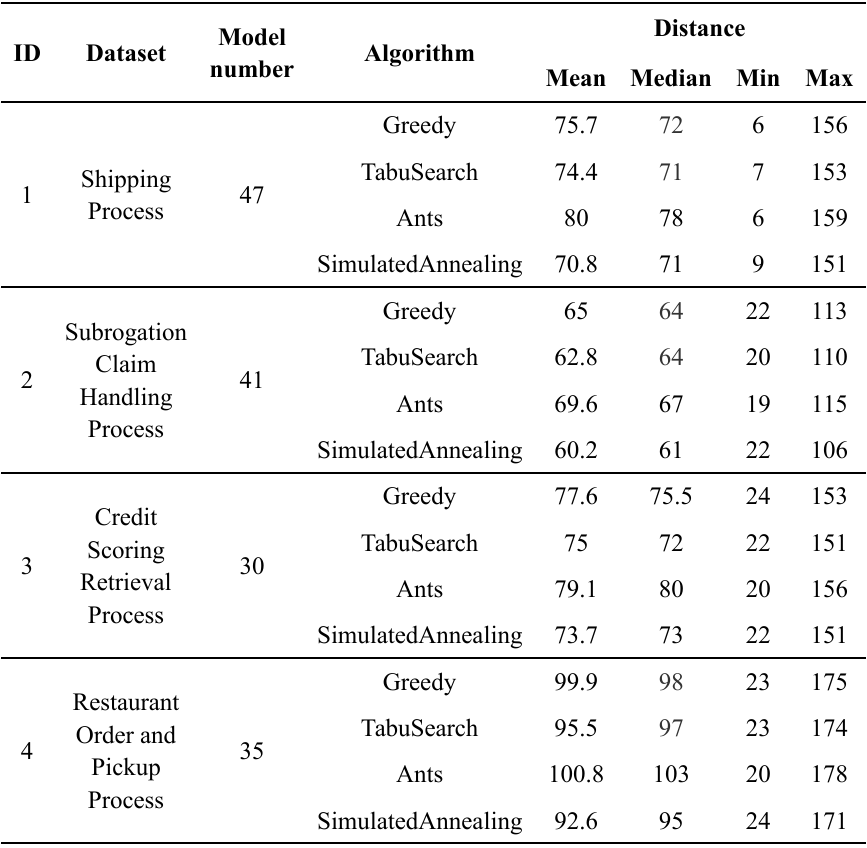}}
  \end{minipage}
\end{tabular}
\end{table}
\;\;\;\;\textbf{\textit{Datasets:}} To evaluate the performance of MAO for generating process models, we choose two types of datasets for comparison, fine-grained and coarse-grained. The fine-grained dataset is sourced from Camunda, a business process management company, we name this dataset as FG-C\footnote{\url{https://github.com/camunda/bpmn-for-research}}, while the coarse-grained dataset is from the Object Management Group (OMG), named CG-O\footnote{\url{https://www.omg.org/spec/BPMN/2.0}}. A detailed description of the two datasets is given below:

\textbf{- FG-C.} In this type dataset, user modeling requirements are described in text form in great detail, including the business objectives to be addressed, the activities involved, and the execution logic between activities. FG-C is collected by Camunda during their BPMN training sessions. Camunda first provides BPMN modeling training to participants, and at the end of the training chapters, participants complete four BPMN modeling exercises. Each exercise is presented as a text-based process scenario, and participants are required to create a BPMN model based on this textual process description. Additionally, for each process description, Camunda provides a reference BPMN model, which participants can study after completing their modeling. In the experiment, we regard this reference BPMN as the standard model. Therefore, the dataset includes four process description texts, each corresponding to multiple BPMN processes modeled by the training participants and one standard BPMN process. As shown in Table \ref{tab:e_table1}, the process description texts in FG-C cover multiple fields, including shipment, insurance claims, financial services, and catering services. Each process description text was modeled by 30-50 participants. To assess the participants' BPMN modeling proficiency and accuracy, we use the BPMNDiffViz tool \cite{ivanov2015bpmndiffviz} with four algorithms to calculate the differences between the BPMN models created by the participants and the standard model for each process description text. In addition, we show the average, median, minimum, and maximum distances between the manually created BPMN models and the standard model in Table \ref{tab:e_table1}.

\textbf{- CG-O.} In this type dataset, user modeling requirements are described very crudely, only including the modeling objectives without any other details of the process, such as generating a purchasing process.  CG-O is released by OMG after the publication of ``The Business Process Model and Notation Specification Version 2.0'' to assist in interpreting and implementing various aspects of the BPMN 2.0 specification. CG-O consists of a collection of process names paired with their corresponding BPMN models. Hence, we consider process names as user requirements and BPMN models as the standard answers. These processes encompass various domains such as transportation, food catering services, and order execution.

\textbf{\textit{Baselines:}} In recent years, research on large language models has made significant progress and garnered considerable attention. However, the study of utilizing large language models for automating the generation of process models is still in its early stages. To our knowledge, only the ProMoAI method proposed by Kourani et al. \cite{kourani2024process} bears similarity to the work presented in this paper. Therefore, we choose the ProMoAI and manual modeling methods as baselines in this paper to evaluate the performance of MAO.

 \textbf{\textit{Metrics:}} To quantify the dissimilarity between the process models obtained through LLM-based modeling methods and the standard answers, we utilize BPMNDiffViz as a comparative tool to evaluate the distance between the process models derived from LLMs and the standard answers. BPMNDiffViz supports multiple algorithms to measure the distance between two process models. In this paper, we selected the Greedy, TabuSearch, Ants, and Simulated Annealing algorithms for distance computation as these algorithms have shown to maintain high accuracy \cite{skobtsov2019efficient}. Table \ref{tab:e_table1} lists the distance results between manual modeling and standard answers calculated by these four algorithms on FG-C.

\subsection{Comparison of experiments on the dataset FG-C}
\begin{figure*}[t]
\centering
  \subfigure[Distance calculation based on Greedy algorithm] { \label{tong1}
  \includegraphics[scale = 0.48]{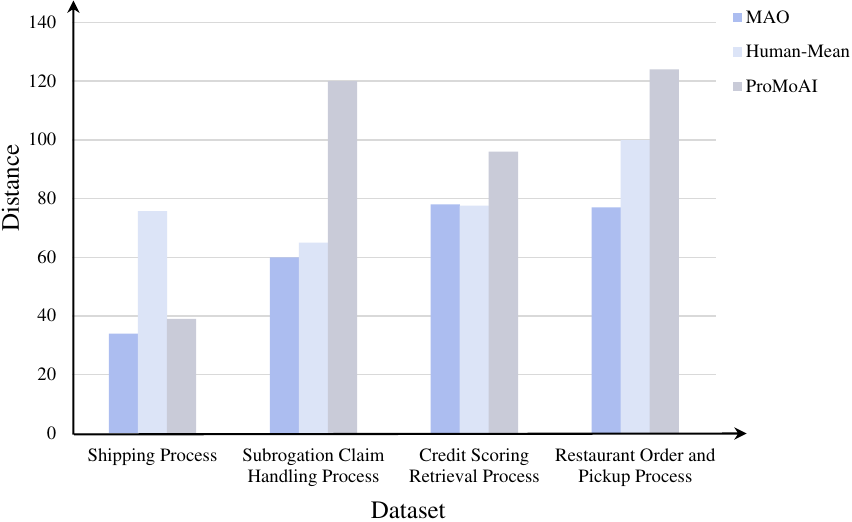}
  }
  \subfigure[Distance calculation based on TabuSearch algorithm] { \label{tong2}
  \includegraphics[scale = 0.48]{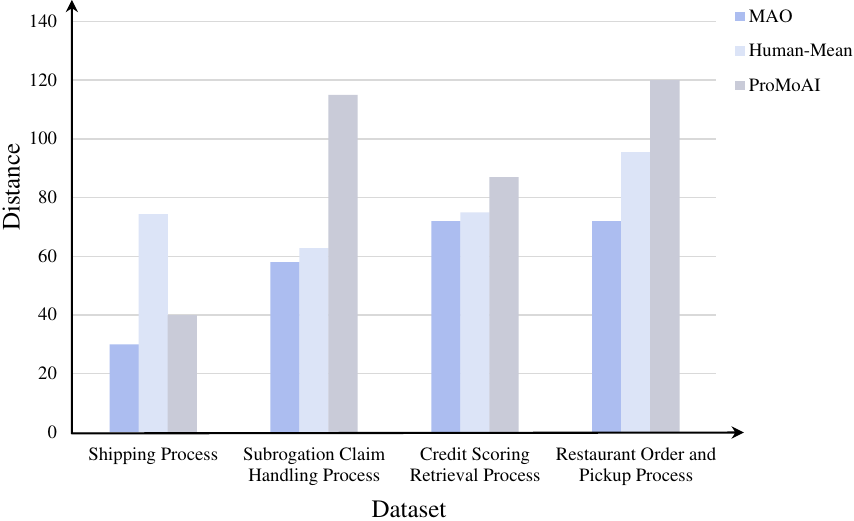}
  }
  \subfigure[Distance calculation based on Ants algorithm] { \label{tong3}
  \includegraphics[scale = 0.48]{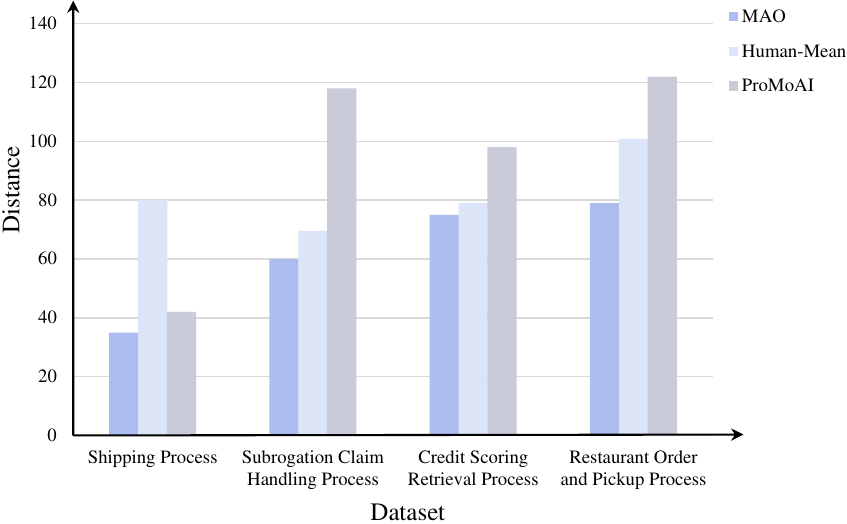}
  }
  \subfigure[Distance calculation based on SimulatedAnnealing algorithm] { \label{tong4}
  \includegraphics[scale = 0.48]{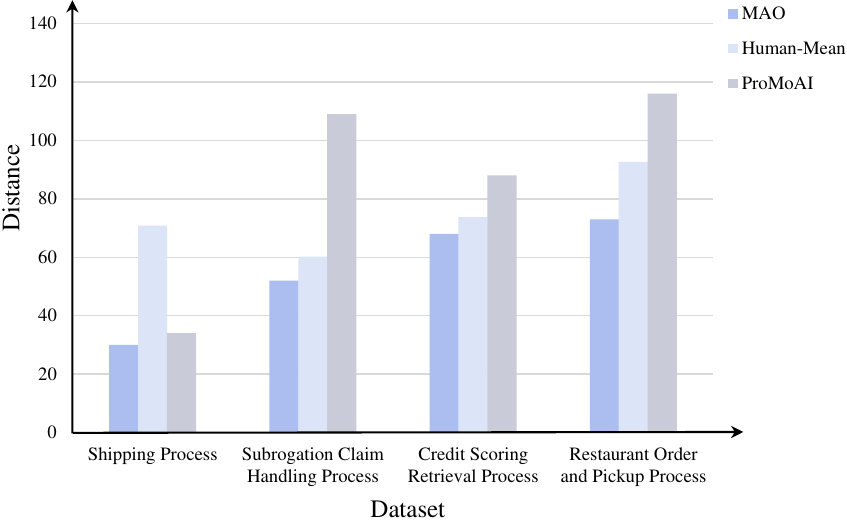}
  }
\caption{Distance between the process models generated by the three methods and the standard models.}
\label{fig:tong}
\end{figure*}

In this experiment, we compare the MAO with the ProMoAI and manual modeling method. We utilize the process description texts from the FG-C provided by Camunda as inputs. The MAO and ProMoAI methods are sequentially applied to generate process models, producing their respective outputs. For the manual modeling method, BPMN models created by multiple participants in Table \ref{tab:e_table1} are considered as outputs. It should be noted that for manual modeling methods, we use the average value of each dataset, which we named Human-Mean.

Next, we employ four algorithms from the BPMNDiffViz tool to compute the distance between the process models generated by the three methods and the standard models. A smaller distance value indicates a closer resemblance of the generated model to the standard answer, hence smaller distance reflects higher model quality. As depicted in Figure \ref{fig:tong}, the four subgraphs represent the distance values computed by different algorithms. Overall, the process models derived from the MAO framework exhibit superior performance compared to ProMoAI and Human-Mean. Specifically, on the ``Shipping Process'' dataset, both MAO and ProMoAI methods significantly outperform manual modeling due to the simplistic gateway structure and limited activity count in the standard model. On the ``Subrogation Claim Handling Process'' dataset, it can be found that ProMoAI is much worse than MAO and Human-Mean, because the gateway structure of the standard model is very complex, and it is difficult for ProMoAI to accurately capture the relationship between activities. On the ``Credit Scoring Retrieval Process'' dataset, the MAO and Human-Mean methods perform comparably, with MAO slightly trailing behind Human in Figure \ref{tong1}, yet outperforming Human-Mean in the other three subgraphs. The essential reason for this is that the standard model of this dataset contains a large amount of messaging activity, i.e. message flow, which is not well understood by the MAO approach. Similarly, ProMoAI and Human-Mean do not cope well with this phenomenon. On the final dataset, although the standard model comprises a multitude of activities, the relationships between activities are straightforward, resulting in both the MAO and Human-Mean methods outperforming ProMoAI.

Finally, we want to evaluate the number of human-generated models that can be surpassed by process models based on a large language model on the FG-C dataset. As shown in Table \ref{tab:e_table2}, the four datasets each consist of 47, 41, 30, and 35 manually created process models, where the average values from the four distance algorithms are utilized as the benchmark for each dataset. We can see that the performance of MAO exceeds that of manual modeling across all four datasets. In contrast, the performance of ProMoAI falls below that of manual modeling on all datasets except the first. Particularly in the second dataset, the presence of complex gateway structures in the standard model results in a significant gap between the models generated by ProMoAI and the standard model, with all 41 human-generated models surpassing ProMoAI. In the third dataset, due to the high number of message flows, the MAO method marginally outperforms 52\% of the human-generated models. However, in the first dataset, the MAO method surpasses 89\% of the human-generated models. In other words, if the process models only contain exclusive gateways, parallel gateways, inclusive gateways, activities, and sequence flows, our framework MAO based on multi-agent orchestration can effectively generate process models from user requirements.

\begin{table}[t]
  \caption{Proportion of MAO and ProMoAI surpassing manual modeling.}
  \label{tab:e_table2}
  \begin{tabular}{ccl}
    \begin{minipage}[h]{0.96\columnwidth}
    \centering
    {\includegraphics[width=\textwidth]{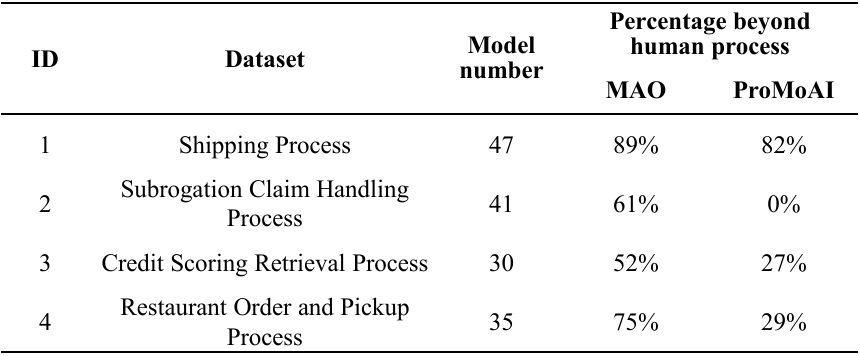}}
  \end{minipage}
\end{tabular}
\end{table}

\subsection{Comparison of experiments on the dataset CG-O}

\begin{figure*}[t]
\centering
  \subfigure[The standard model from CG-O] { \label{BPMN1}
  \includegraphics[width=0.92\textwidth]{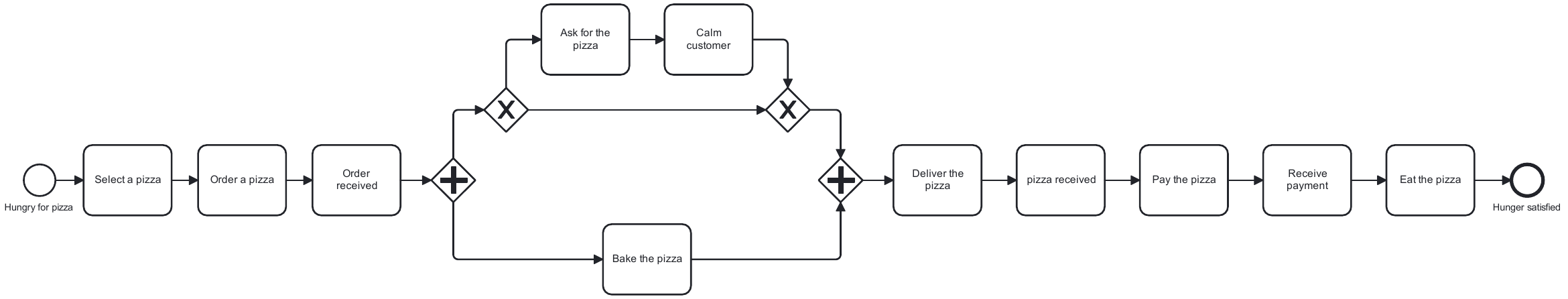}
  }
  \subfigure[Process model generated by MAO] { \label{BPMN2}
  \includegraphics[width=0.92\textwidth]{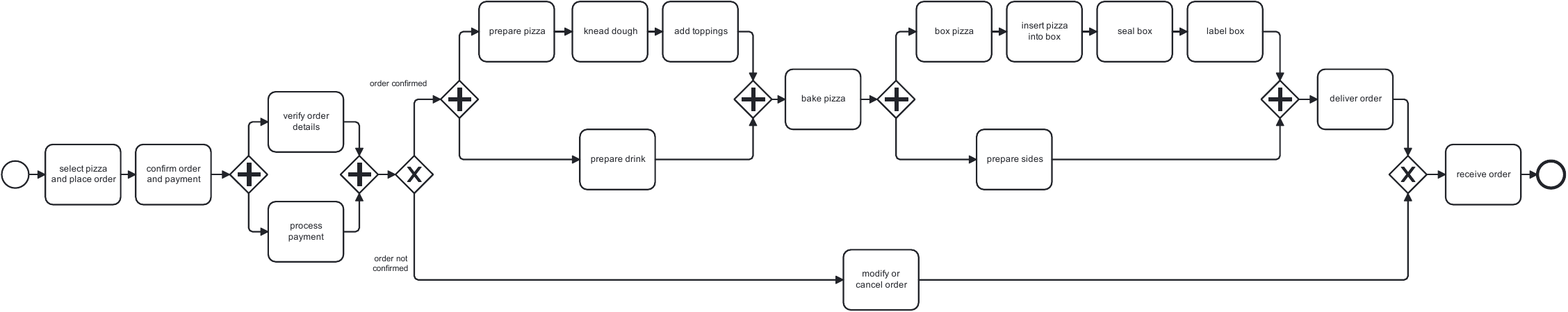}
  }
  \subfigure[Process model generated by ProMoAI] { \label{BPMN3}
  \includegraphics[width=0.69\textwidth]{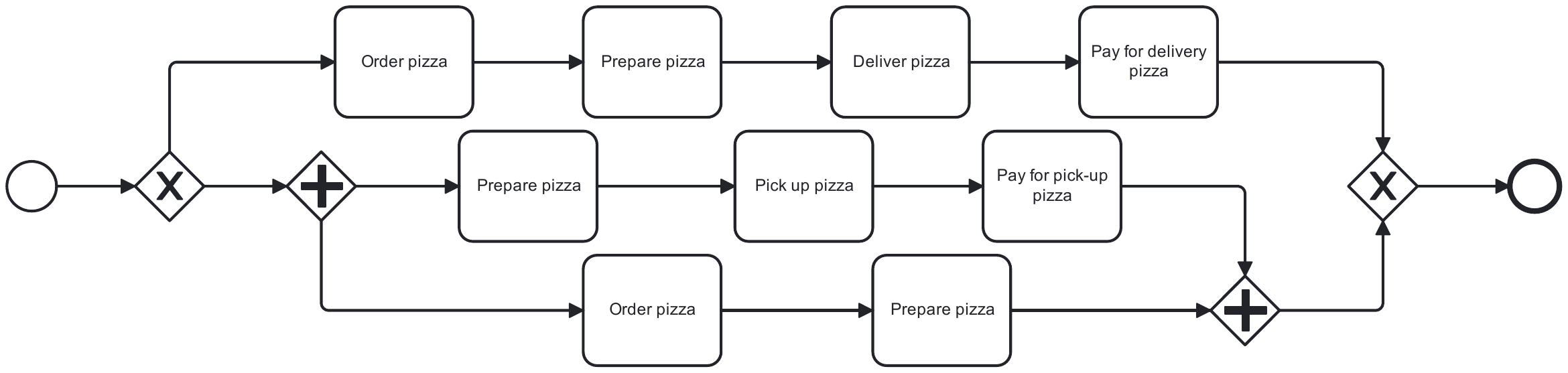}
  }
\caption{Process models generated by different methods on CG-O about ``Ordering and delivering pizza''.}
\label{fig:BPMN}
\end{figure*}

In this section, we want to discuss the capability of large language models in generating processes on open-ended requirements. As stated in Section 4.1, CG-O is a coarse-grained dataset containing cases of three business processes: ``Ordering and delivering pizza'', ``Shipment of a hardware retailer'', and ``Order Fulfillment''. Each case includes only a standard process model and its corresponding name, without any description of the process model. This type of dataset CG-O meets the open-ended requirements, because even with the same business process name, individuals will create different process models based on their own knowledge and experiences.

As depicted in Figure \ref{BPMN1}, in the case of ``Ordering and delivering pizza'', the OMG organization provides a standard model that they consider to represent the business process of ordering pizza in a reasonable manner. Figures \ref{BPMN2} and \ref{BPMN3} illustrate the process models generated by the MAO and ProMoAI based on the user requirement (i.e., ``please design a process model about ordering and delivering pizza''). In the user requirements, there are no descriptions of activities related to ordering or delivering pizza, so both methods create process models based on their own knowledge autonomously. Since open-ended requirements do not have definitive answers, a comparative analysis according to the metric outlined in Section 4.2 is not feasible. Therefore, this paper subjectively evaluates the strengths and weaknesses of the two methods (MAO and ProMoAI). From Figure \ref{BPMN2}, it is apparent that the MAO method generates a more detailed process model compared to the standard model, with all activities within the model being rational. This provides valuable reference for modelers. In contrast, the process model generated by ProMoAI is very rough and may exhibit unreasonable duplicate activities, such as ``Prepare pizza'' appearing three times and ``Order pizza'' appearing twice. We attribute the strong performance of MAO on open datasets to its ability to produce detailed and valuable process models in the first two phases and ensure model quality through hallucinations handling in the subsequent phases. The modeling results for the other two cases in the CG-O dataset are similar to the ``Ordering and delivering pizza'' case and will not be reiterated here. In conclusion, the framework MAO proposed in this paper produces process models of higher quality on open-ended requirements compared to existing methods, offering valuable references for modelers.

With respect to both time efficiency and cost, the performance of MAO also outperforms ProMoAI. The average time for MAO to generate a process model is about 10 minutes, which costs \$0.12. In contrast, building a model with ProMoAI takes 15 minutes and costs \$0.40.

\subsection{Ablation study}
To assess the contributions of each phase in our proposed framework MAO, this section will conduct ablation experiments: 1) Refinement-$\phi$, which represents discarding the second phase in MAO. 2) Reviewing-$\phi$, which removes the Reviewing phase in MAO, generating processes without examining semantic errors in the process text. 3) Testing-$\phi$, which excludes the Testing phase, generating process texts without checking for format errors. 4) MAO, which includes all phase.

\begin{table}
  \caption{Ablation analysis in different datasets, where the values represent the distances from the standard models.}
  \label{tab:e_table3}
  \begin{tabular}{ccl}
    \begin{minipage}[h]{0.96\columnwidth}
    \centering
    {\includegraphics[width=\textwidth]{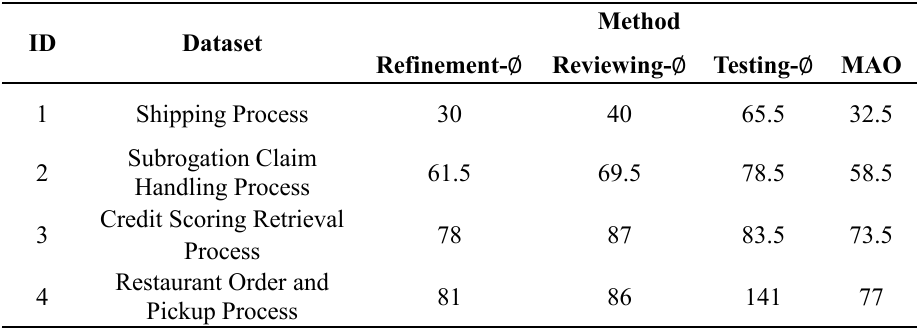}}
  \end{minipage}
\end{tabular}
\end{table}

The findings from the experimental results presented in Table \ref{tab:e_table3} lead to the following conclusions: a) the Refinement phase contributes minimally on fine-grained datasets. Furthermore, if the standard model only includes a small number of activities, the refinement operation results in an increased distance between the generated model and the standard model. For example, in the ``Shipping Process'' dataset, the distance value of Refinement-$\phi$ is lower than that of the MAO; b) if process models consisting solely of sequence flows, the contribution of the Testing phase surpasses that of the Reviewing phase in MAO. For instance, in the ``Restaurant Order and Pickup Process'' dataset, the distance value obtained from Testing-$\phi$ is twice that of MAO, indicating that the Testing phase results in a 50\% reduction in the distance value of MAO. In contrast, the value obtained from Reviewing-$\phi$ is 1.1 times the value of MAO, representing a 11\% reduction in MAO's distance value with the Reviewing phase. Similar conclusions can be drawn from the ``Shipping Process'' and ``Subrogation Claim Handling Process'' datasets; c) if process models involving complex structures like message flows, the contribution of the Reviewing phase outweighs that of the Testing phase in MAO. For example, in the ``Credit Scoring Retrieval Process'' dataset, the distance value obtained from Reviewing-$\phi$ exceeds that obtained from Testing-$\phi$. In summary, both the Reviewing and Testing phases in MAO effectively address hallucinations phenomena in process model generation, leading to process models that closely align with standard models.

\section{Related work}
This study mainly explores process modeling and large language models, with particular emphasis on leveraging multi-agent to automatically produce process models and investigating approaches to deal with the phenomenon of hallucinations in modeling. Consequently, this section will clarify relevant work from the perspectives of traditional process modeling, automated modeling, and hallucinations handling.

\textbf{\textit{Traditional process modeling:}} Process model has always been an indispensable tool \cite{scacchi2002process} in the field of software engineering, as it can assist developers in quickly capturing user requirements \cite{udousoro2020effective}, depicting software components to support software reengineering \cite{dogru2003process}, and simulating software systems to optimize software development \cite{garcia2020software}. Commonly used modeling languages include BPMN \cite{object2011business}, UML \cite{booch1997uml}, Petri Net \cite{van2013business}, EPC \cite{scheer2005process}, and YAWL \cite{van2005yawl}. Each modeling language has its specifications and characteristics \cite{pereira2016business}. Against this background, De Leoni et al. \cite{de2021integrating} provide a native, formal definition of DBPMN models, which integrates BPMN and DMN standards. They also provide their formal execution semantics through an encoding into Data Petri Nets (DPN), demonstrating how to use them. However, traditional process modeling has been typically carried out by modeling experts or business personnel, rather than being automated. Even though RPA workflows can execute processes automatically, the processes still necessitate human intelligence for elaborate design \cite{wewerka2020robotic}. The drawback of manual modeling is that it can be time-consuming due to factors such as personal emotions or busy work schedules. Furthermore, the involvement of multiple domain experts or business personnel can escalate modeling costs.

\textbf{\textit{Automated modeling:}} Process mining is a cross-disciplinary field at the intersection of data mining and process modeling, enabling the automated construction of process models from the execution logs of information systems \cite{kalenkova2016process, van2018process}. Process modeling techniques based on process mining are so efficient that the entire automated creation process often takes only a few minutes. However, the only datasets that earlier process discovery algorithms can handle are event logs consisting of sequences of activities \cite{van2004workflow, leemans2014discovering}. For instance, in a simple procurement business process, the event log is expressed as $\mathcal{L}$ = [$\sigma_0$ = ⟨a, b, d⟩, $\sigma_1$ = ⟨a, b, d⟩, $\sigma_2$ = ⟨a, b, d⟩,  $\sigma_3$ = ⟨a, c, d⟩, $\sigma_4$ = ⟨a, c, d⟩, $\sigma_5$ = ⟨a, b, d⟩, $\sigma_6$ = ⟨a, c, d⟩], where a trace $\sigma$ represents a complete execution record of the business process, with activities a, b, c and d denoting ``Purchase materials'', ``Alipay'', ``Pay Pal'' and ``Receive materials'', respectively. Friedrich et al. \cite{friedrich2011process} achieved process model generation from text datasets by specifying mining rules. The BPMN Sketch Miner \cite{ivanchikj2020text} utilizes process mining to derive BPMN models from textual representations in a domain-specific language. Qian et al. \cite{qian2020approach} combined natural language processing techniques to extract process models from process descriptive text, transforming the task of process modeling into a multi-grained text classification problem. They efficiently extracted multi-grained information using a hierarchical neural network, thereby constructing BPMN models. Nevertheless, existing text-based process modeling technologies necessitate detailed activity information within text data, posing limitations when dealing with open-ended user requirements.

In recent years, with the rapid development of large language models (LLMs), their performance in addressing open-ended questions has been widely recognized, such as generating images from textual descriptions with DALL \cite{ramesh2021zero}, generating code based on user requirements with AlphaCode \cite{li2022competition}, and Codegen \cite{nijkamp2022codegen}. Consequently, the integration of LLMs in process modeling has been investigated recently \cite{busch2023just, vidgof2023large}. Ye et al. \cite{ye2023proagent} proposed ProAgent using LLM-based agents, which can achieve workflow construction and dynamic decision-making in workflow execution. However, ProAgent fails to consider hallucination phenomena. Zeng et al. \cite{zeng2023flowmind} utilized large language models to create an automatic workflow generation system, named Flowmind. However, the workflows generated by Flowmind exhibit simple sequential structures and struggle to handle complex structures like concurrency and selection. In \cite{kourani2024process}, the ProMoAI framework can generate process models containing complex structures from process description text. However, addressing hallucination issues in the process requires human intervention, and the quality of generated process models relies on the intermediate process representation (i.e., POWL).

\textbf{\textit{Hallucinations handling:}} Hallucination can be a problem in NLP as it would lead to misleading or incorrect information, which can be categorized into factuality hallucination and faithfulness hallucination \cite{huang2023survey}. Similarly, hallucination phenomena can arise in process modeling. But the existing modeling approaches based on LLMs rely on human intervention to mitigate model errors. As we know, various automated methods exist to reduce hallucination in LLMs. For instance, Zhang et al. \cite{zhang2023alleviating} propose a straightforward Induce-then-Contrast Decoding (ICD) strategy to mitigate hallucinations. Retriever-Augmented Generation (RAG) is also a promising solution to address hallucinations \cite{gao2023retrieval}, which synergistically integrates the intrinsic knowledge of LLMs with dynamic external databases to enhance the LLM's domain knowledge. Furthermore, Martino et al. \cite{martino2023knowledge} leverage knowledge injection to enhance large language models performance at a relatively low cost, thereby increasing enterprise users' confidence in utilizing LLMs and enabling them to opt for smaller, more cost-effective LLMs. In addition to the aforementioned approaches for handling hallucinations, Lei et al. \cite{lei2023chain} have also explored the use of Chain of Natural Language Inference (CoNLI) for hallucination detection. 

\vspace{-0.2cm}
\section{conclusion}
Process modeling is a valuable topic in software engineering. This paper explores a novel process modeling framework, named MAO. The MAO framework leverages the capabilities of large language models to create a process design team composed of multiple agents, and then finishes process modeling with multi-agent orchestration. Upon receiving user requirements, the MAO framework goes through four phases: Generation, Refinement, Reviewing, and Testing to produce a BPMN diagram. The first two phases involve designing a rough process model and refining the process model, while the latter two phases deal with semantic hallucinations and format hallucinations. Extensive experiments were conducted on two types of datasets, fine-grained dataset FG-C and coarse-grained dataset CG-O. The results indicate that for user requirements containing detailed process descriptions, the process model quality generated by the MAO surpasses the existing methods and far exceeds the average level of manual modeling. For open-ended requirements, i.e., those containing only modeling objectives, the MAO framework is capable of generating a reasonable process model with significant reference value.

In the future, we will expand the work content of this article to support more different elements of BPMN in process models, such as data flow and message flow.


\bibliographystyle{ACM-Reference-Format}
\bibliography{main}

\end{document}